\documentclass[twoside]{article}
\usepackage{multicol,headrule,ams2la}
\usepackage{graphicx}
\usepackage{bm}
\usepackage{amssymb}
\usepackage{float}
\usepackage{multirow}
\usepackage{array}
\usepackage{booktabs}
\usepackage{caption}
\usepackage{ulem}
\usepackage{amsbsy}
\usepackage{makecell}
\usepackage{tabularx}
\usepackage{color}
\usepackage{ulem}

\usepackage{ragged2e} 
\usepackage{booktabs}

\graphicspath{{figs/}}
\newcolumntype{L}[1]{>{\raggedright\arraybackslash}p{#1}}
\newcolumntype{C}[1]{>{\centering\arraybackslash}p{#1}}
\newcolumntype{R}[1]{>{\raggedleft\arraybackslash}p{#1}}

\graphicspath{{figs/}}
\renewcommand{\arraystretch}{1.03}

\def\oddhead{{\rm\hfill\protect\small
Fine-grained Cross-modal Fusion based Refinement for Text-to-Image Synthesis
\hfill }}
\def\evenhead{{\rm \hfill \protect\small\it Chinese Journal of
Electronics  \rm\hfill 2023}}

\def\RE{\par\vskip1.5\kh\par\centerline{\bf
References}\par\vskip0.5\kh\par}
\def\REF#1{\par\hangindent\parindent\indent\llap{#1\enspace}\ignorespaces}

\DeclareSymbolFont{lettersA}{U}{txmia}{m}{it}
\DeclareMathSymbol{\piup}{\mathord}{lettersA}{25}
\DeclareMathSymbol{\muup}{\mathord}{lettersA}{22}

\begin{document}
\newtheorem{theorem}{Theorem}
\newtheorem{proposition}{Proposition}
\newtheorem{remark}{Remark}
\newtheorem{proof}{Proof}
\def\footnoterule{\kern 1mm \hrule width 12cm \kern 2mm}
\abovedisplayskip=8.0pt plus 2.0pt minus 1.5pt
\belowdisplayskip=8.0pt plus 2.0pt minus 1.5pt
 \def\thefootnote{}
\TagsOnRight
\def\kh{\baselineskip}
\def\pmb#1{\boldsymbol{#1}}

\thispagestyle{empty}
\noindent Chinese Journal of Electronics


\vskip 2cm

\begin{center}\huge\bf

Fine-grained Cross-modal Fusion based Refinement for Text-to-Image Synthesis
\end{center}

\vskip 11mm

\begin{center}\large\rm
SUN Haoran$^{2}$, WANG Yang$^{1, 2, *}$, LIU Haipeng$^{2}$, QIAN Biao$^{2}$
\\\vskip1mm
{\small (1.~\it  Key Laboratory of Knowledge Engineering With Big Data, Ministry of Education, Hefei University of Technology, \\ Hefei, China
\rm)}

{\small (2.~\it Department of Computer Science and Information Engineering, Hefei University of Technology, Hefei, China
\rm)}


\footnotetext{\footnotesize * \ WANG Yang is the corresponding author (Email: yangwang@hfut.edu.cn) }
\end{center}

\begin{multicols}{2}

{\footnotesize\bf Abstract --- {\footnotesize\bf
Text-to-image synthesis refers to generating visual-realistic and semantically consistent images from given textual descriptions. Previous approaches generate an initial low-resolution image and then refine it to be high-resolution. Despite the remarkable progress, these methods are limited in fully utilizing the given texts and could generate text-mismatched images, especially when the text description is complex. We propose a novel Fine-grained text-image Fusion based Generative Adversarial Networks, dubbed FF-GAN, which consists of two modules: Fine-grained text-image Fusion Block (FF-Block) and Global Semantic Refinement (GSR). The proposed FF-Block integrates an attention block and several convolution layers to effectively fuse the fine-grained word-context features into the corresponding visual features, in which the text information is fully used to refine the initial image with more details.
And the GSR is proposed to improve the global semantic consistency between linguistic and visual features during the refinement process. Extensive experiments on CUB-200 and COCO datasets demonstrate the superiority of FF-GAN over other state-of-the-art approaches in generating images with semantic consistency to the given texts.Code is available at https://github.com/haoranhfut/FF-GAN.}

}

\begin{center}{\large\bf I.\ Introduction
}\end{center}

Text-to-image synthesis is one of the most significant tasks in the field of natural language processing$^{[24, 26, 42]}$ and computer vision$^{[8, 14, 19, 25,29,39,41]}$, which aims to synthesize visual-realistic and text-matched images from the given linguistic descriptions. With the recent success of the Generative Adversarial Networks (GANs)$^{[3, 12, 16, 25, 29, 31, 32]}$, text-to-image synthesis has drawn increasing attention and a great number of advanced methods$^{[2, 4, 18, 27, 33]}$ have been proposed.

Most approaches adopt a fashion of multi-stage generation$^{[4, 5, 8, 10, 29, 33, 35]}$ to obtain high-quality images,
which first generates a coarse image by utilizing sentence-level textual feature and improve it to be high-resolution. Although conventional approaches are impressive in generating high-quality images, most of these approaches often synthesize images that mismatch with the given text semantics and fail to fully utilize the text information, in particular when the text is complicated. One major reason for this problem is the ineffective and inadequate fusion of text and image information during the refinement process of these methods.
Early works$^{[4,5,13,27,28,40]}$ attempt to simply concatenate the encoded text information with the visual feature or utilize the attention mechanism$^{[30, 37, 41]}$ to integrate the cross-modal features. However, the semantic gap between different modalities seriously impedes the fusion of texts and images. For example, Attn-GAN$^{[5]}$ employs an attention mechanism to fuse the fine-grained word-level linguistic and visual information, which first utilizes cross-modal attention to obtain the word-context features for each image sub-region. And then concatenates the word-context features with the corresponding image features to refine the initial image. However, simply concatenating features of two different modalities is sub-optimal because it cannot explicitly distinguish which regions to be refined. Recently, SD-GAN$^{[34]}$ adopts conditional batch normalization$^{[1, 15, 17, 43, 45]}$ to inject text information into the image feature map. However, conducting batch normalization on the visual feature map and transforming it into a normal distribution may cause the loss of visual diversity.

\vskip -3mm
\begin{figure}[H]
\includegraphics[width=0.49\textwidth]{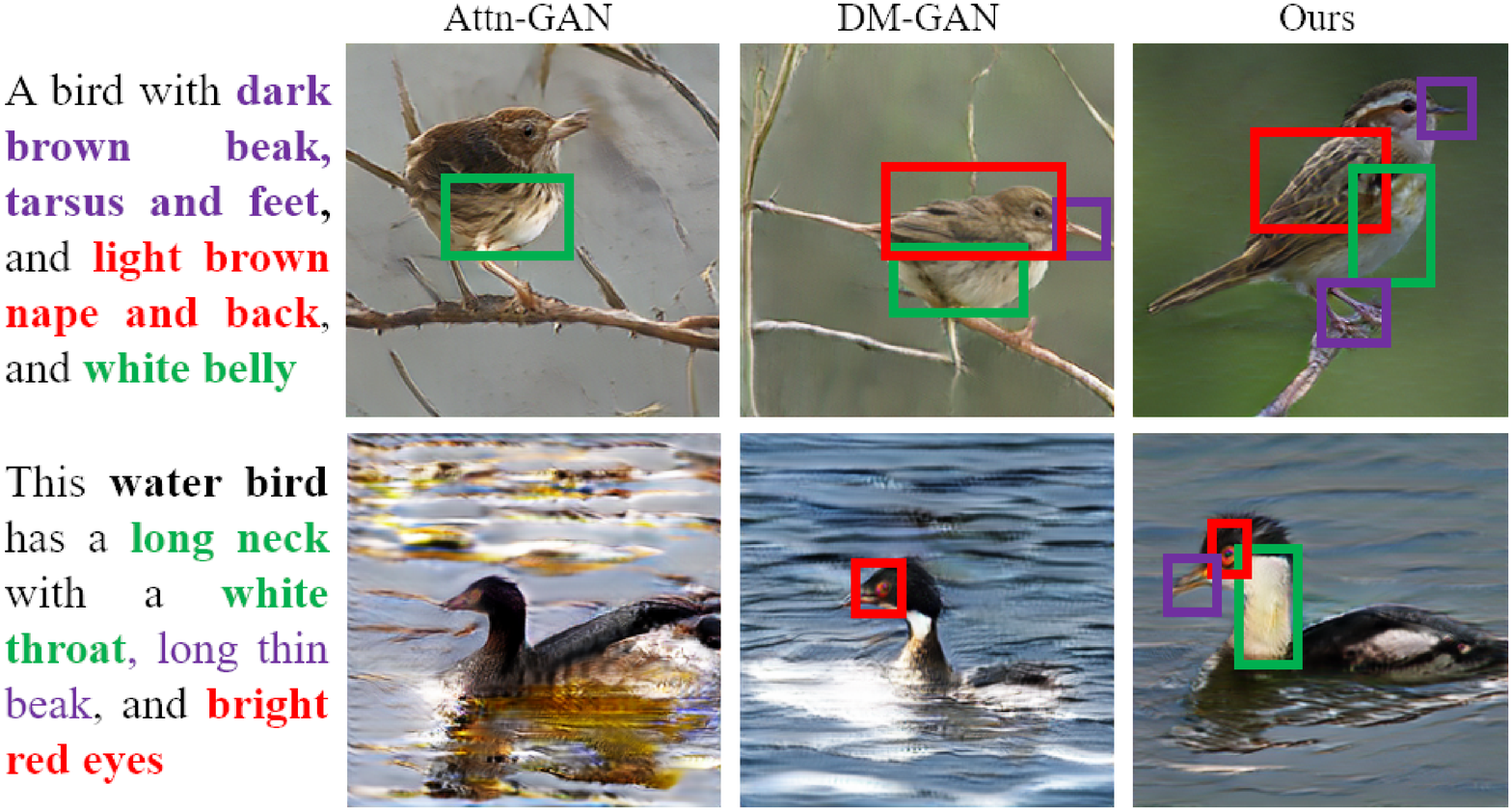}
\label{fig1}
\end{figure}
\vskip -6mm
\centerline{\footnotesize\begin{tabular}{l} Fig.\ 1.\ \ Samples generated by Attn-GAN$^{[5]}$, DM-GAN$^{[35]}$ and our \\  model on CUB-200 dataset. Images generated by our method  are \\ more  realistic and  semantically consistent with the given texts.
\end{tabular}}
\vskip 2mm

By fusing cross-modal features, fine-grained word-level linguistic features can be used to refine the visual features, which can make full use of text information while adding more vivid details to the initial images. However, previous methods cannot fully fuse the cross-modal features, and thus synthesize images inconsistent with the given text descriptions. As shown in Fig.1, images generated by Attn-GAN$^{[5]}$ and DM-GAN$^{[35]}$ failed to match the texts semantically. In order to synthesize more text-matched images, we need to effectively inject fine-grained linguistic features into visual features to refine the initial images. Also, the global semantics is required to encourage the generated images semantically consistent with the texts.

To this end, we develop a novel Fine-grained text-image Fusion based Generative Adversarial Networks (FF-GAN) to improve the quality of images synthesized from conditional textual descriptions.
Specifically, the first mechanism is the Fine-grained text-image Fusion Block (FF-Block), which selects more important word features that contains fine-grained linguistic information and conducts affine transformation on the visual feature maps. In this way, fine-grained textual information can be added to its corresponding image sub-region in a sufficient and efficient manner. The second novel mechanism is Global Semantic Refinement (GSR), which introduces a global semantic constraint during the refinement phase. Combing the GSR with the FF-Block in the refinement process makes it gradually and smoothly drives the generator toward the goal of fine-grained semantic alignment both globally and locally.
We performed extensive experiments on two widely used and challenging
benchmark datasets to verify the performance of the FF-GAN in the task of
text-to-image synthesis. Our FF-GAN shows remarkable superiority compared with the most advanced approaches in two evaluation metrics. Overall, the main contributions of this paper can be summarized as follows:

1. We develop a novel end-to-end framework named FF-GAN, which makes full use of
given textual descriptions to produce more visual-realistic and text-matched images.

2. An effective FF-Block is proposed to fuse cross-modal features more adequately and efficiently, and a GSR is developed to improve the global semantic alignment during the refinement phase.

\begin{center}{\large\bf II.\ Related Work
}\end{center}

{\bf 1.\ GAN for text-to-image synthesis}

The success of GAN has greatly promoted the development of text-to-image generation, and many models based on GANs and its variants have been proposed. For the first time, Reeds et al.$^{[2, 11]}$ use conditional GAN$^{[12, 18]}$ to generate $64 \times 64$ fuzzy resolution images from texts. In a bid to improve the quality of synthesized pictures, Stack-GAN $^{[4]}$ is
proposed, which synthesizes visual-authentic images with two stages by stacking a series of generators and discriminators. First, a coarse image is generated in the low-stage, and then the details of the image are modified to generate a high-resolution
one. Attn-GAN$^{[5]}$ introduces an attention mechanism to align the semantics of texts and images. By combining semantically aligned image and text features, the generated images may maintain semantic consistency with the texts. DM-GAN$^{[35]}$ develops a dynamic memory based network to obtain the more significant linguistic feature relevant to the synthesized images at each image refinement process. DAE-GAN$^{[46]}$introduces extra knowledge named aspect information to improve the details of synthesized images both locally and globally during the attention-guided generation process.

 The aforementioned methods proposed many impressive mechanisms to align the feature of different modalities, which have achieved remarkable success in the field of text-image generation. However, simply concatenating aligned cross-modal representations is sub-optimal for the fusion of cross-modal features, which leads to insufficient use of text semantics to generate semantic-consistent images.


In order to fuse the text and visual features more adequate, SD-GAN$^{[34]}$ adopts conditional batch normalization to inject text information into the visual feature maps instead of concatenating the aligned cross-modal features directly, which is more sufficient for the fusion of cross-modal features. However, the normalization of visual features will lead to the loss of visual information diversity, which is very bad for our text-image generation task.

Different from the previous methods which simply splice different aligned modal features, or utilize CBN to correlate text and image representation, we propose to use affine transformation to fuse the features of different modalities. In this way, we introduce a more adequate and efficient word-level text-image fusion that enables the model to generate fine-grained images with high quality.

{\bf 2.\ Affine transformation}

Affine transformation is widely used in the case of conditional batch normalization to introduce additional conditional information$^{[1, 15, 17]}$ and avoid the loss of information$^{[43, 45]}$ caused by normalization.

\end{multicols}
\vskip -4mm
\begin{figure}[H]
\centerline{\includegraphics[width=0.99\textwidth]{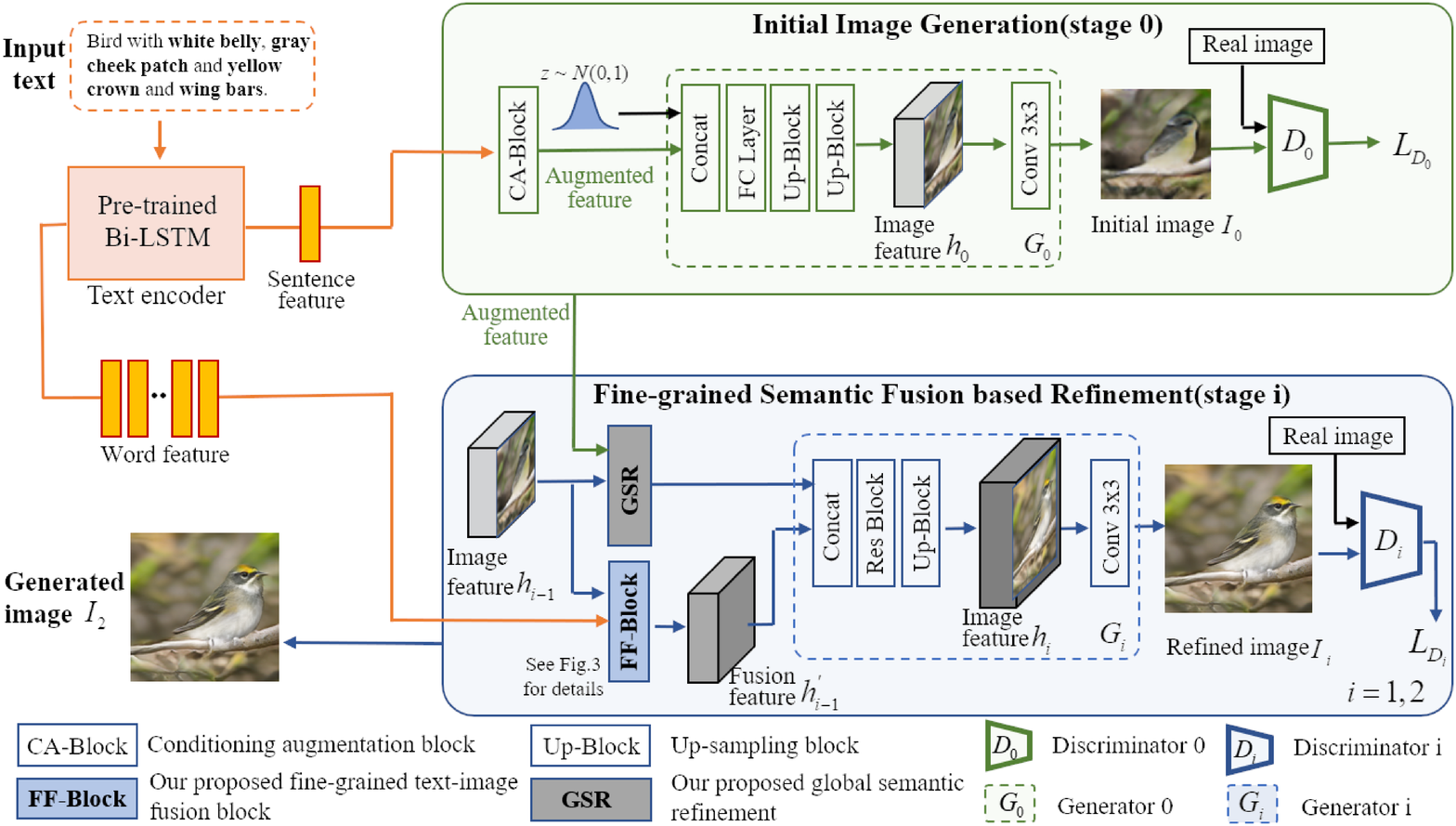}}
\end{figure}
\vskip -2mm
\centerline{\footnotesize\begin{tabular}{l} Fig.\ 2.\ Framework of our proposed FF-GAN. Our FF-GAN utilizes Bi-LSTM to extract linguistic features of two granularities, i.e., the \\ sentence-level and word-level features, then it generates a low-resolution image by using the sentence-level features in the initial image \\ generation (stage 0) and refines it to obtain a high-resolution image in fine-grained semantic fusion based refinement by using  both the\\word-level features and sentence-level features (stage 1 and stage 2).

\end{tabular}}
\vskip -2mm
\begin{multicols}{2}

Mani-GAN$^{[47]}$ introduces affine transformation in semantic image manipulation. It uses an affine transformation to fuse the cross-modal representations between texts and images for effectively editing the image.
MS-GAN$^{[10]}$ introduces an attention-modulation block to fuse the representations of text and image. DF-GAN$^{[44]}$ also proposed a deep text-image fusion block that stacks multiple affine transformations to make a full text-image fusion.

Although the aforementioned approaches have achieved remarkable results in the fusion of cross-modal features, only fusing sentence-level semantics and visual feature map will lead to affine transformation work on the visual representation in a spatially uniform manner and result in inadequate fusion. Ideally, the linguistic information should be incorporated into the sub-region of images with corresponding semantics.
Different from the above methods, our proposed model introduces more fine-grained word-level affine transformation to fully integrate the cross-modal features, which reinforces the generative model to synthesize authentic and text-matched images with more vivid details.

\begin{center}{\large\bf III.\ Proposed Method
}\end{center}

As Fig.2 shows, the framework of the proposed FF-GAN mainly consists of two modules: the initial image generation and fine-grained semantic fusion based refinement.  In initial image generation, we first extract semantic representations from the given textual description into sentence-level and word-level features, then we synthesize a initial low-resolution image according to the sentence-level linguistic feature. In the refinement stage, we develop an effective Fine-grained text-image Fusion Block (FF-Block) which fuses word-level features into visual features to refine the initial image with details, while a Global Semantic Refinement (GSR) to improve the globally semantic consistency, in a bid to synthesize high-quality images that match with the corresponding textual sentences.

{\bf 3.1\ Initial image generation}

We employ a bidirectional LSTM$^{[24, 26, 42]}$ to encode the input semantic representation into two granularity, namely sentence-level feature and the word-level feature. The sentence-level text feature $S$ is leveraged to synthesize the initial image in this phase and the word-level representation $W$ is utilized to refine the initial image in the following refinement phase.
\begin{equation}
\label{f8}
\{W,S \} = LSTM(T), \\
\end{equation}
where $T = \{T_t | t = 0, \cdots , L - 1\} $ is the text description which contains $L$ words, $W = \{ W_t | t = 0, 1, \cdots, L - 1\} \in \mathbb{R}^{D_w\times L}$ is the word-level feature by concatenating the hidden states of LSTM, $S \in \mathbb{R}^{D_w}$ is the sentence-level feature from the last hidden layer of LSTM, and $D_w$ is dimension of $W_t$ and $S$.

Limited training data could lead to the sparsity in the textual conditioning manifold, so we follow Stack-GAN$^{[4]}$ by using Conditioning Augmentation (CA) to augment input text information. It yields more training data and thus improves the robustness of model against small perturbations. Explicitly, we enhance sentence-level linguistic feature $S$ with conditioning augment function $F_{ca}$ and obtain the augmented sentence semantic representation $S_{ca}$ as follows:
\begin{equation}
\label{f8}
S_{ca} = F_{ca} (S).
\end{equation}

Then, we employ the augmented linguistic feature $S_{ca}$ and a noise vector $z \sim N (0, 1)$ that is stochastically sampled from a normal distribution to synthesize an initial image. Formally, we can obtain the initial image $I_0 \in \mathbb{R}^{D_m \times D_m}$ and its corresponding image feature $h_0 \in \mathbb{R}^{D_m \times D_m}$ as follows:
\begin{equation}
\label{f8}
\{I_0, h_0\} = G_0 (S_{ca}, z),
\end{equation}
where $G_0$ is the generator which is composed of a fully connected layer and multiple up-sampling layers.

\vskip 1mm
\begin{figure}[H]
\includegraphics[width=0.48\textwidth]{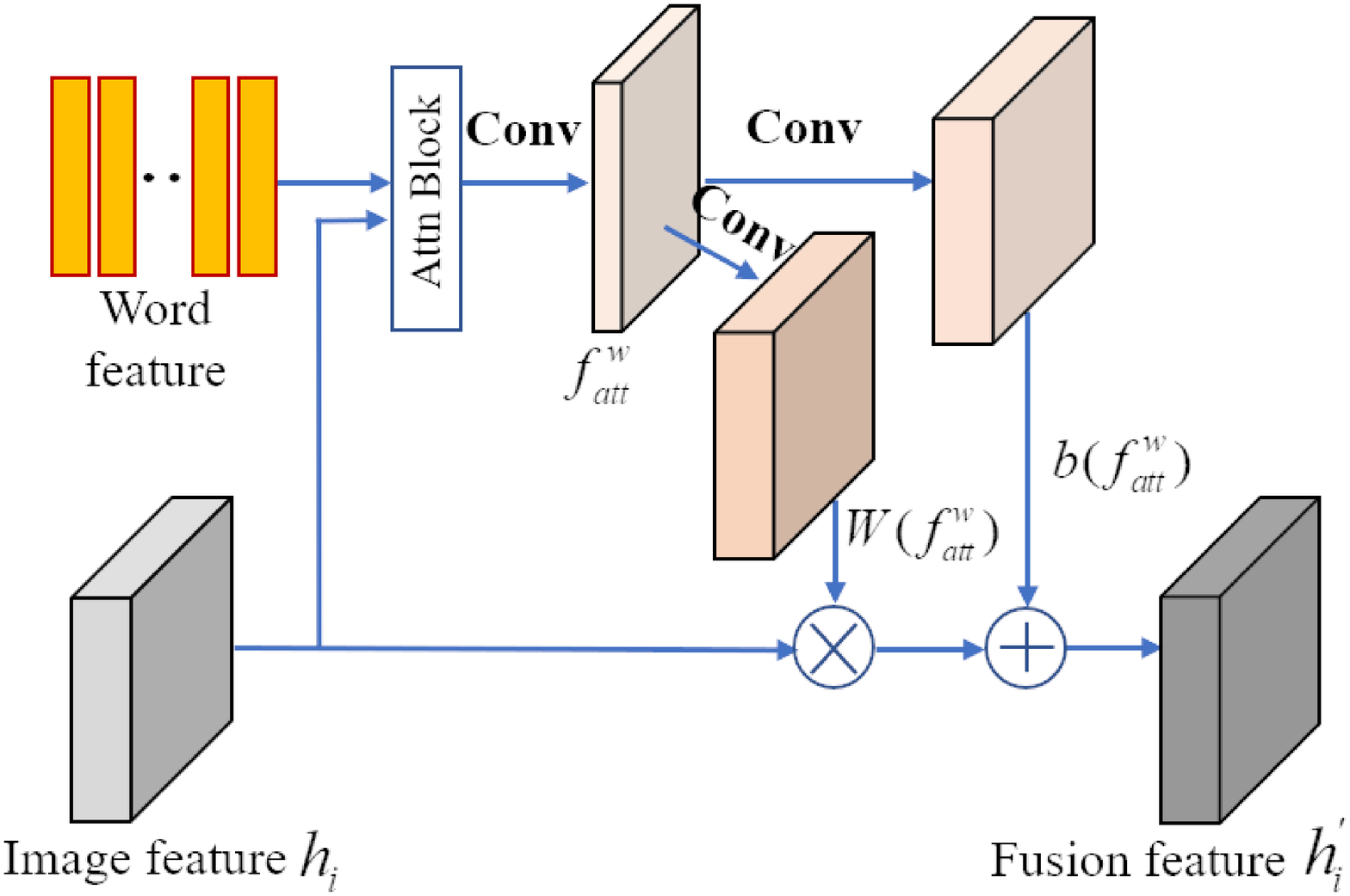}
\end{figure}
\vskip -2mm
\centerline{\footnotesize\begin{tabular}{l} Fig.\ 3.\ The architecture of our proposed FF-Block, which aligns \\ the word feature and image feature  by an attention block, and \\ utilizes the affine transformation to fuse text and image features  \\ effectively.
\end{tabular}}
\vskip 2mm

{\bf 3.2\ Fine-grained semantic fusion based refinement}

The image generated in the initial generation stage is coarse with only a rough shape and always ignores some important details of the text description. In this stage, we propose an FF-Block to fuse word-level features into visual feature maps, which contributes to refining the initial images with more fine-grain linguistic information. Also, We develop a GSR to improve the globally semantic consistency of cross-modal features.

{\bf 3.2.1\ Fine-grained semantic-aware fusion}

In this section, we propose a novel FF-Block, as can be seen from Fig.3, which contains an attention block and several convolution layers, to fuse the fine-grained word-level feature into the image feature maps for refining the initial images with more vivid details.

Each word in the textual description possesses different importance, so it is sub-optimal to fuse the word-level linguistic feature and image feature directly. We employ an attention mechanism to obtain an attentive word-context feature, which contributes to determining the importance of every word for cross-modal fusion.
Specifically, we firstly utilize a preception layer $U_w$ to map the word-level linguistic feature $W_i$ into the identical latent space of visual feature.
Then we compute an attention score between the word-level feature $W_i$ and image feature $h_0$ by the softmax function. We obtain the attentive word-context linguistic feature $f_{att}^{W}$ by conducting the inner product between $U_w W_i$ and the attention score as follows:
\begin{equation}
\label{f8}
f_{att}^{w} = \sum_{i=0}^{L-1}  (U_w W_i) (softmax (h_0^\mathrm{T}(U_w W_i)))^\mathrm{T},
\end{equation}
where $U_w \in \mathbb{R}^{D_m \times D_w}$ and $f_{att}^{w} \in \mathbb{R}^{D_m \times D_m} $, and the dimension of attentive word-context feature $f_{att}^{w}$  is the same as visual feature $h_0$.

To integrate linguistic and visual features efficiently, we utilize the attentive word-context feature $f_{att}^{w}$ to conduct affine transformation on the visual feature $h_0$. To be specific, we adopt several convolution layers to process the attentive word-context feature $f_{att}^{w}$ and then predict the linguistic-conditional channel-wise scaling matrix $W (f_{att}^{w})$ and shifting matrix  $b (f_{att}^{w})$ that both have the same size as $h_0$. Finally, we obtain a cross-modal fusion representation $h_{0}^{ \prime }$ by fusing the word-context feature $f_{att}^{w}$ and image feature  $h_0$ as follows:
\begin{equation}
\label{f8}
h_{0}^{ \prime } =  h_0 \odot W (f_{att}^{w}) + b (f_{att}^{w}),
\end{equation}

where $W (f_{att}^{w})$ and $b (f_{att}^{w})$ are the learned weights and bias term based on the attentive word-context feature $f_{att}^{w}$, and $\odot$ is the Hadamard element-wise product.
Our FF-Block is able to effectively integrate word-level text features and image features to achieve a fine-grained cross-modal fusion. We could comprehend the effect of the affine transformation in cross-modal fusion from the following two aspects:

 (1) scaling matrix $W (f_{att}^{w})$: By multiplying with the scaling term $W (f_{att}^{w})$, it helps establish the relationship between image feature $h_0$ and linguistic feature $f_{att}^{w}$, while re-weight the visual feature $h_0$. Sub-regions of image feature $h_0$ that match with the word-context feature $f_{att}^{w}$ will be precisely highlighted, otherwise it will be weakened, which plays a role in regional selection. In this way, FF-Block can accurately identify the attributes in the image that match the word-context semantics and establish the correlation between attributes and word-context semantics, so as to encourage the refined image of better image-text consistency.

 (2) shifting matrix  $b (f_{att}^{w})$: The bias term $b (f_{att}^{w})$ can effectively encode the text information and introduce the details in text descriptions which are ignored by initial images. Meanwhile, as mentioned in Feat-Trans$^{[6]}$, the bias term can achieve similar effects as the implementation of concatenation, which contributes to taking full advantage of linguistic information while retaining the invariant information in the visual features.

By conducting the above affine transformations, the fine-grained word-context features can be efficiently fused into the visual feature maps to realize the fine-grained modification of the initial coarse images, and the refined images will contain more vivid details and be better consistent with the textual descriptions.

{\bf 3.2.2\ Global semantic refinement}

We have introduced an FF-Block to effectively fuse cross-modal features to refine the initial images in the previous part. However, such a local fusion may lead to an inconsistency between the synthesized images and the linguistic semantics from the global semantic perspective. Some words with a higher importance in the descriptions may even affect the direction of the whole image refinement process, while those relatively less important word-context features may be ignored.

In order to maintain the global semantic consistency, we introduce the GSR which exploits the sentence-level attention mechanism.
The augmented linguistic feature $S_{ca}$ is first mapped into the same latent space as the visual feature by a perceptron layer $U_s \in \mathbb{R}^{D_m \times D_w}$. Then it conducts the softmax function on the visual feature to attain the similarity score between the cross-modal features. Specifically, we could obtain a sentence-context feature $f_{att}^{S}$ as follows:
\begin{equation}
\label{f8}
f_{att}^{S} =  (U_s S_{ca}) (softmax (h_0^\mathrm{T}(U_s S_{ca})))^\mathrm{T}.
\end{equation}


To this end, the sentence-context feature $f_{att}^{S}$ will be concatenated with cross-modal fusion representation $h_{0}^{ \prime }$ and then fed into the $i_{th} (i = 1, 2)$ generator $G_i$ to synthesize a refined high-resolution image $I_i$ and its corresponding visual feature $h_i$ as follows:
\begin{equation}
\label{f8}
\{I_i, h_i\} = G_i (h_{0}^{ \prime }, f_{att}^{S}),
\end{equation}
where the generator $G_i$ integrates a series of up-sampling layers and a residual block.

It is worthy to note that the combination of sentence-context features and word-level fusion representations can not only ensure the consistency of visual-linguistic semantics, but also help smoothen the refinement progress from both global and local perspectives.

{\bf 3.3\ Objective functions}

To generate visual-realistic images and maintain the cross-modal semantic consistency, we design the final objective function of generators which contains two practical adversarial losses:  the unconditional loss and the conditional loss. To be specific, the loss function of the generator $G_{i}(i=0,1,2)$ can be expressed as follows:
\begin{equation}
\label{f8}
L_{G_i} = \underbrace{-\frac{1}{2}\mathbb{E}_{x \sim p_{G_i}}[log (D_i (x))]}_{unconditional\ loss}
        \underbrace{-\frac{1}{2}\mathbb{E}_{x \sim p_{G_i}}[log (D_i (x, S_{ca}))]}_{conditional\ loss},
\end{equation}
where $x$ is the generated image sampled from the distribution $p_{G_i}$ in the $i_{th}$ stage, and $S_{ca}$ is the augmented linguistic features. The first part is an unconditional loss that estimates the input image to be visually authentic or fake, while another one is the conditional loss, which is utilized to discriminate whether the image and input text is a correct match.

Following the common practice$^{[5, 33, 35]}$, we further adopt the practical DAMSM$^{[5]}$ loss to compute the alignment degree between linguistic and visual features, denoted as $L_{DAMSM}$. To smoothen the conditional manifold and avoid over-fitting$^{[4]}$, we also employ a regularization term $L_{CA}$ during training,  which is defined as the Kullback-Leibler divergence  (KL divergence) between the standard Gaussian distribution and the conditioning Gaussian distribution of the texts. Mathematically, the $L_{CA}$ is defined as follows:
\begin{equation}
\label{f8}
L_{CA} = D_{KL} (\mathcal{N} (\mu (S_{ca}), \Sigma (S_{ca}))||\mathcal{N} (0,1)),
\end{equation}
Where $\mu (S_{ca})$ is the mean and $\Sigma (S_{ca})$ is the diagonal covariance matrix of the linguistic feature, which are computed by a perception layer.
Finally, we define the final loss function of the generative neural networks as follows:
\begin{equation}
\label{f8}
L_{G} = \sum_{i}L_{G_i} + \lambda_1 L_{CA} + \lambda_2 L_{DAMSM}.
\end{equation}

Similarly, the adversarial loss of discriminator $D_{i}$ also contains an unconditional loss and a conditional loss. Mathematically, the loss function of discriminator $D_{i}$ can be defined as:
\begin{small}
\begin{equation}
\begin{aligned}
\label{f8}
L_{D_i} &= \underbrace{-\frac{1}{2}[\mathbb{E}_{\hat{x}\sim p_{data}}[log (D_i (\hat{x}))] + \mathbb{E}_{x\sim p_{G_i}}[log (1-D_i (x))]}_{unconditional\ loss} \\
        &+ \underbrace{\mathbb{E}_{\hat{x}\sim p_{data}}[log (D_i (\hat{x}, S_{ca}))] + \mathbb{E}_{x\sim p_{G_i}}[log (1-D_i (x, S_{ca}))]]}_{conditional\ loss},
\end{aligned}
\end{equation}
\end{small}

where the $\hat{x}$ is the image sample from the realistic image distribution $p_{data}$. The final objective function of the discriminator is defined as follows:

\begin{equation}
\label{f8}
L_D = \sum_iL_{D_i}.
\end{equation}

\begin{center}{\large\bf IV.\ Experiments
}\end{center}

In this section, we performed extensive experiments on the CUB-200$^{[23]}$ and COCO$^{[22]}$ benchmark datasets to evaluate the proposed FF-GAN. Section 4.1 details the datasets, evaluation metrics, and training details used in the experiments. Then we compare our FF-GAN quantitatively and qualitatively with the advanced GAN-based methods, e.g., Attn-GAN$^{[5]}$, Mirror-GAN$^{[33]}$, DM-GAN$^{[35]}$, et al. We also perform extensive ablation experiments on the key components of our proposed FF-GAN.

{\bf4.1\ Experiment setup}

{\bf4.1.1\ Datasets}

We perform the experiments on two challenging open datasets, i.e., CUB-200$^{[23]}$ and COCO$^{[22]}$. The CUB-200 contains 200 species of  bird with about 12k images. Each bird image is annotated with ten linguistic descriptions. The training set of CUB-200 includes 9k images of 150 bird categories while the test set has 50 categories with 3k images. The COCO dataset consists of 80k training images and 40k testing images. Each image is annotated with five textual sentences in the COCO dataset.

{\bf4.1.2 Evaluation metrics}

Following the practice, we quantify the performance of our method and related competitors on two widely used quantitative metrics: ${\rm Fr\acute{e}chet}$ Inception Distance  (FID)$^{[21]}$  and R-precision$^{[5]}$. FID evaluates the realism of the synthetic images by computing the ${\rm Fr\acute{e}chet}$ distance between the visual feature distribution of the synthesized and authentic images, which are extracted by a Inception-V3 network$^{[37]}$ pre-trained by Attn-GAN${[5]}$. A lower FID score suggests a higher realism of the generated images.
Following Attn-GAN$^{[5]}$, we use R-precision to evaluate whether the synthesized pictures are well-conditioned on the corresponding textual sentences. Specifically, we compute the cosine similarities between a query visual feature and 100 candidate linguistic features extracted from R matched text descriptions and 100-R stochastically selected descriptions in the dataset. Then we rank the results and retrieve the top-R matched sentences to get the R-precision score. In practice, we set R$=$1 in our experiment. Higher R-precision values suggests that the synthetic image is much semantically consistent with the corresponding textual description.

To compute the FID score and R-precision score, we randomly select textual captions from the test set to generate 30000 images from each model, with each image of 256 $\times$ 256 resolution.

{\bf 4.1.3\ Implementation details}

Our proposed FF-GAN first generates a $64 \times 64$ image in the initial generation stage, then the fine-grained semantic fusion based refinement stage refines the initial image to $128 \times 128$ and $256 \times 256$ resolution. Note that we only repeat the refinement process twice for the GPU memory limitation. Following DM-GAN$^{[35]}$, we apply spectral normalization for all discriminators to enhance the stability of the training process and improve performance. We utilize a pre-trained bidirectional LSTM$^{[24,25,26]}$ by Attn-GAN$^{[5]}$ to produce the sentence-level and word-level representations. Meanwhile, we set the dimension of word-level vector to 256, the dimension of the augmented sentence vector is 100 and the sentence length is 18 for the balance of performance and calculation.
Following the Attn-GAN$^{[5]}$ and DM-GAN$^{[35]}$, we set the hyper-parameter $\lambda_1$ to 1, $\lambda_2 $ is 5 on CUB-200 and 50 on COCO respectively.
We adopt the Adam optimizer with a learning rate set to 0.0002 during training on one Geforce GTX 1080Ti GPU.
Then, we train our proposed FF-GAN with 600 epochs on CUB-200 and 120 epochs on COCO.

{\tabcolsep=1.5pt \footnotesize
\begin{center}
\renewcommand{\arraystretch}{1.5}
\begin{tabular}{|C{2.7cm}|C{2.7cm}|C{2.7cm}|}
    \multicolumn{3}{c}{\makecell[c]{\bf Table. 1. R-precision (higher is better) on the test set of \\ \bf CUB-200  and COCO  (* means using extra supervisions).\\ \bf  The two
        best scores are marked with \textcolor[RGB]{255,0,0}{red} and \textcolor[RGB]{0,0,255}{blue} color.}} \\ \hline
   $\textbf{Method}$   &  $\ \ \ \ \ \textbf{CUB-200}$ \ \ \ \ \  \ \ \ & $ \textbf{COCO} $   \\ \hline
  $\   \mbox{Attn-GAN}^{[5]} \  $& $ 67.82\pm4.43 $  & $ 72.31\pm 0.91 $   \\ \hline
  $\   \mbox{Mirror-GAN}^{[33]} \  $& $ 57.67 $ &  $ 74.52 $  \\ \hline
  $\   \mbox{RiFeGAN}^{[36]} \  $& $ 23.8\pm 1.5 $ &  $ - $   \\ \hline
  $\   \mbox{Control-GAN}^{[9]} \  $& $ 69.33\pm 3.23 $ &  $ 82.43 \pm 2.43 $   \\ \hline
  $\   \mbox{DM-GAN}^{[35]} \  $& $ \textcolor[RGB]{0,0,255}{72.31\pm 0.91}$ &  $ 88.56\pm 0.28 $   \\ \hline
  $\   \mbox{KT-GAN}^{*[38]} \  $& $ 32.9 $ &  $ 24.5 $   \\ \hline
  $\   \mbox{Huang et al.}^{*[20]} \  $& $ - $ &  $ \textcolor[RGB]{0,0,255}{89.69\pm 4.34} $   \\ \hline
  $\   \mbox{TIME}^{*[7]} \  $& $ 71.57\pm 1.2 $ &  $ 89.57\pm 0.9 $   \\ \hline
  $\   \mbox{\textbf{Ours}} \  $& $ \textcolor[RGB]{255,0,0}{80.49 \pm 0.50} $  & $ \textcolor[RGB]{255,0,0}{91.28 \pm 0.51} $   \\ \hline
\end{tabular}
\end{center}}

{\tabcolsep=1.5pt \footnotesize
\begin{center}
\renewcommand{\arraystretch}{1.5}
\begin{tabular}{|C{2.7cm}|C{2.7cm}|C{2.7cm}|}

      \multicolumn{3}{c}{\makecell*[c]{\bf Table. 2.\ FID scores  (lower is better) on the test set  of  \\ \bf CUB-200 and COCO  (* means using extra supervisions). \\ \bf  The two
        best scores are marked with \textcolor[RGB]{255,0,0}{red} and \textcolor[RGB]{0,0,255}{blue} color} }\\ \hline

   $\textbf{Method}$   &  $\   \ \textbf{CUB-200}$  \ & $ \textbf{COCO} $   \\ \hline

  $\   \mbox{HD-GAN}^{[13]} \  $& $ 18.23$ &  $ 75.34  $   \\ \hline
  $\   \mbox{Attn-GAN}^{[5]} \  $& $ 23.98$  & $ 35.49 $   \\ \hline
  $\   \mbox{Mirror-GAN}^{[33]} \  $& $ 18.34$ &  $ 34.71 $  \\ \hline
  $\   \mbox{DM-GAN}^{[35]} \  $& $ 16.09$ &  $ 32.64$  \\ \hline
  $\   \mbox{KT-GAN}^{*[38]} \  $& $ 17.32 $ &  $ \textcolor[RGB]{0,0,255}{30.73}$  \\ \hline
  $\   \mbox{Huang at el.}^{*[20]} \  $& $ - $ &  $ 34.52$  \\ \hline
  $\   \mbox{TIME}^{*[7]} \  $& $ \textcolor[RGB]{255,0,0}{14.30} $ &  $ 31.14 $   \\ \hline
  $\   \mbox{\textbf{Ours}} \  $ & $ \textcolor[RGB]{0,0,255}{15.13}$  & $ \textcolor[RGB]{255,0,0}{29.44}$   \\ \hline

\end{tabular}

\end{center}}

{\bf4.2\ Quantitative results}

We compare our method quantitatively with several advanced methods on the test set of  CUB-200 and COCO, the performance results on two evaluation metrics are reported in Tabel. 1 and Table. 2. It is worth mentioning that recent approaches often employ extra supervisions, e.g., KT-GAN$^{[38]}$ uses the extra teacher network, and TIME$^{[7]}$ uses extra 2-D positional encoding.

It is illustrated that our model achieves competitive performance compared with other remarkable methods in all evaluation metrics, especially in the R-precision score which measures the semantic alignment between the synthesized pictures and the conditioning texts. Our method remarkably increases the R-precision score by a large margin from 72.31 to 80.49 on the CUB-200 as compared with DM-GAN$^{[35]}$, which achieves remarkable success in the task of text-to-image synthesis. As for the more challenging COCO dataset with multiple objects and complex backgrounds, our FF-GAN also outperforms other approaches and achieves a 91.28 R-precision score by a 1.59 numerical improvement compared with Huang et al.$^{[20]}$, and 2.72 numerical improvements compared with DM-GAN$^{[35]}$. The superior performance of our method demonstrates that our FF-GAN is able to synthesize images that are more semantically consistent with the given texts, thanks to the FF-Block which effectively fuses the text and visual features in an efficient manner.

Table. 2 shows the performance on the CUB-200 and COCO datasets with respect to the FID score. As can be seen, the FID score of our FF-GAN is 15.13, which is only inferior to the result given by TIME$^{[7]}$ that uses extra 2-D positional encoding, but much better than other advanced models: 23.98 in Attn-GAN$^{[5]}$, 18.34 in Mirror-GAN$^{[33]}$ and 16.09 in DM-GAN$^{[35]}$. As for the more challenging COCO dataset, our method shows remarkable superiority over all advanced approaches and decreases the FID score to 29.44, which indicates our FF-GAN is able to synthesize high-quality pictures with multiple complex sub-objects in highly realistic. Excellent performance on FID illustrates our FF-GAN can generate more authentic images than other state-of-the-art approaches.


{\bf4.3\ Qualitative results}

{\bf4.3.1\ Subjective visual comparisions}

 In order to qualitatively evaluate the visual quality of our proposed FF-GAN, we compare our qualitative results with the most advanced approaches, i.e., Attn-GAN$^{[5]}$ and DM-GAN$^{[35]}$. As  Fig.4(a) and Fig.4(b) shows, the images synthesized by FF-GAN contain more fine-grained details and are more semantically alignment with the given textual descriptions.

\end{multicols}
\vskip-4mm
\begin{figure}[H]
\centerline{\includegraphics[width=0.98\textwidth]{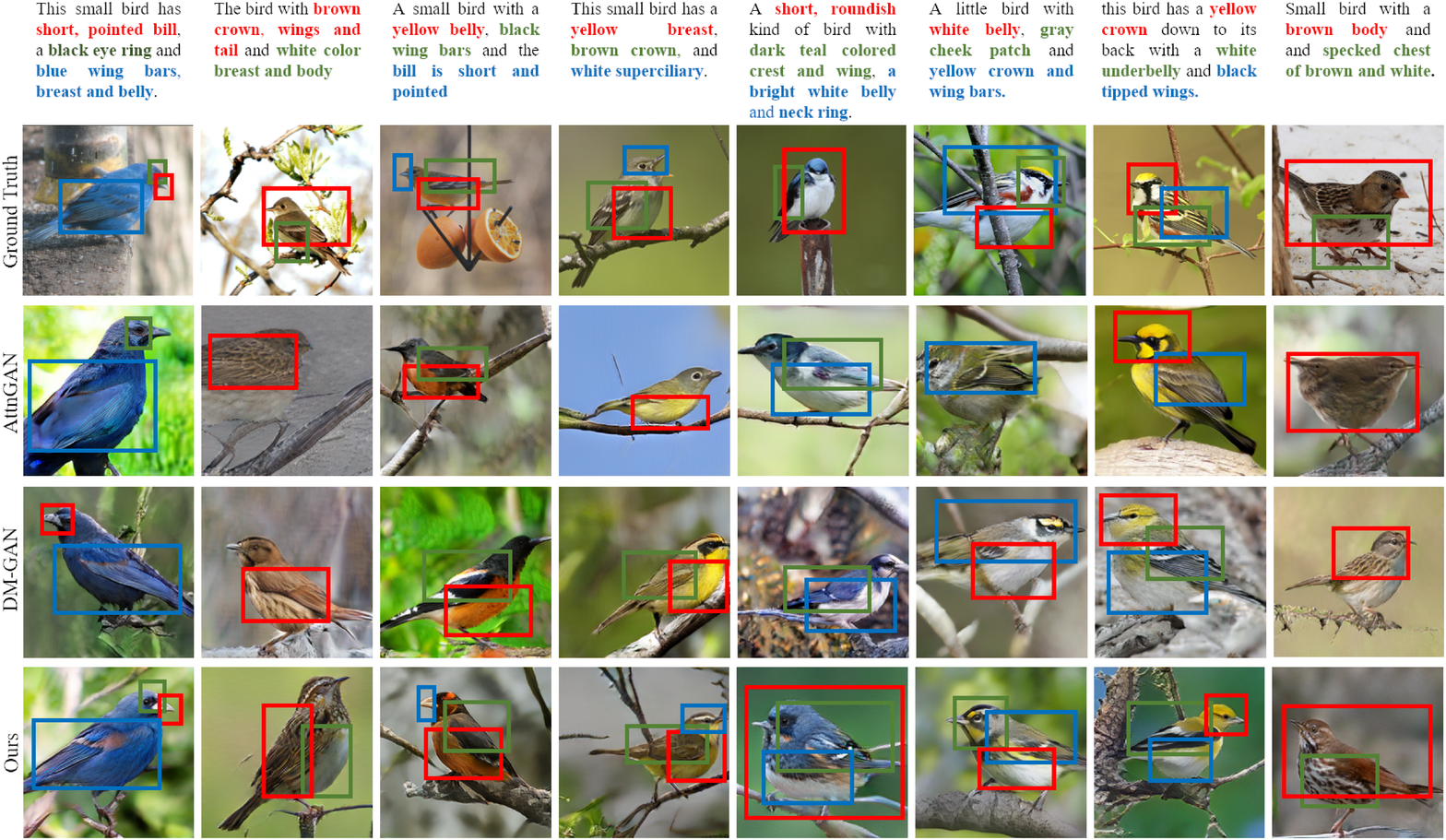}}
\end{figure}
\vskip -1mm
\centerline{\footnotesize\begin{tabular}{c}  (a) The CUB-200 dataset.
\end{tabular}}
\vskip-2mm
\begin{figure}[H]
\centerline{\includegraphics[width=0.98\textwidth]{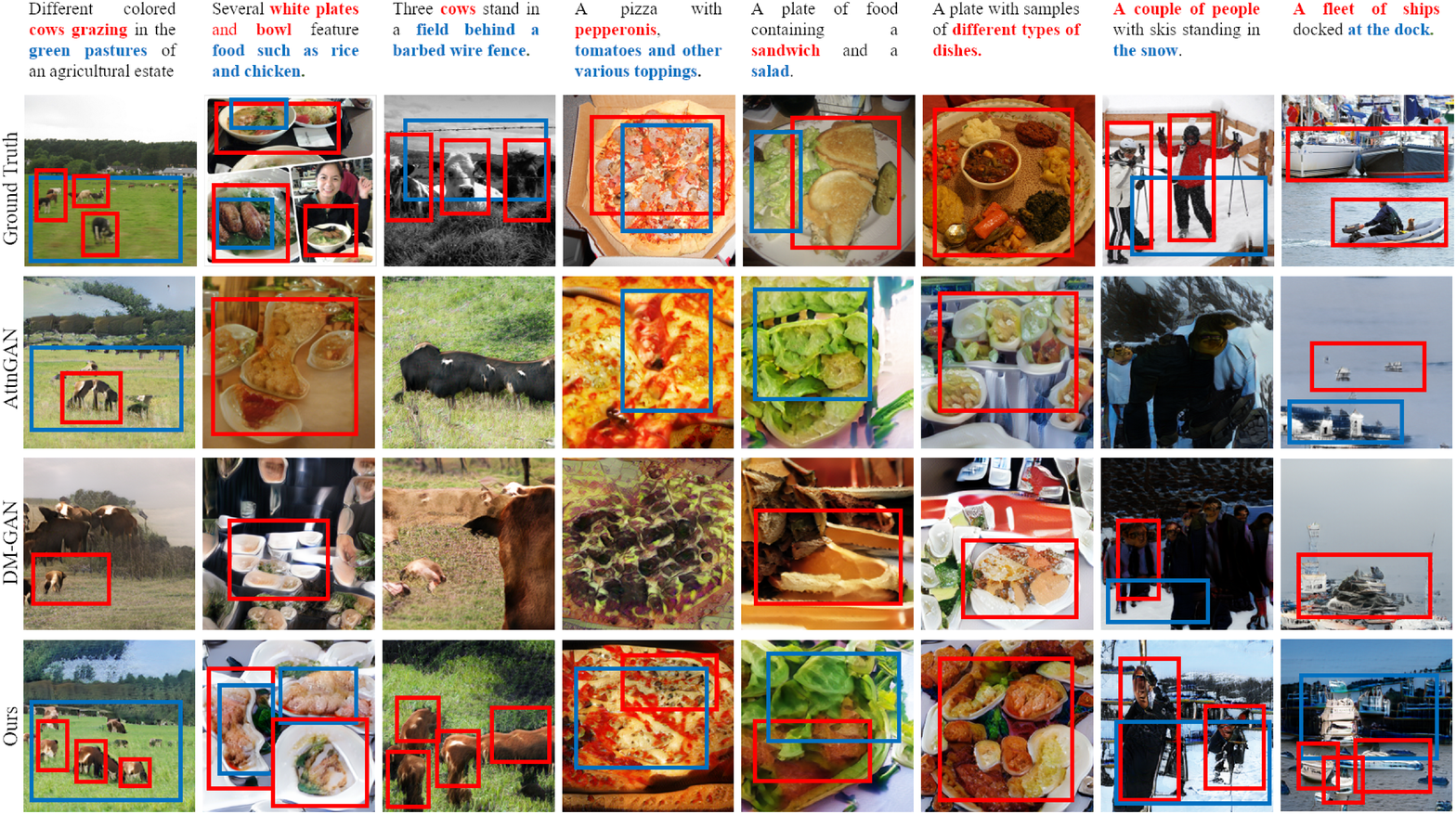}}
\end{figure}
\vskip -1mm
\centerline{\footnotesize\begin{tabular}{c}  (b) The COCO dataset.
\end{tabular}}
\vskip 2mm
\centerline{\footnotesize\begin{tabular}{c} Fig.\ 4.\ Qualitative comparison between our method and advanced Attn-GAN,  DM-GAN on CUB-200  (a) and COCO datasets (b).
\end{tabular}}

\begin{multicols}{2}

\end{multicols}
\vskip-8mm
\begin{figure}[H]
\centerline{\includegraphics[width=0.98\textwidth]{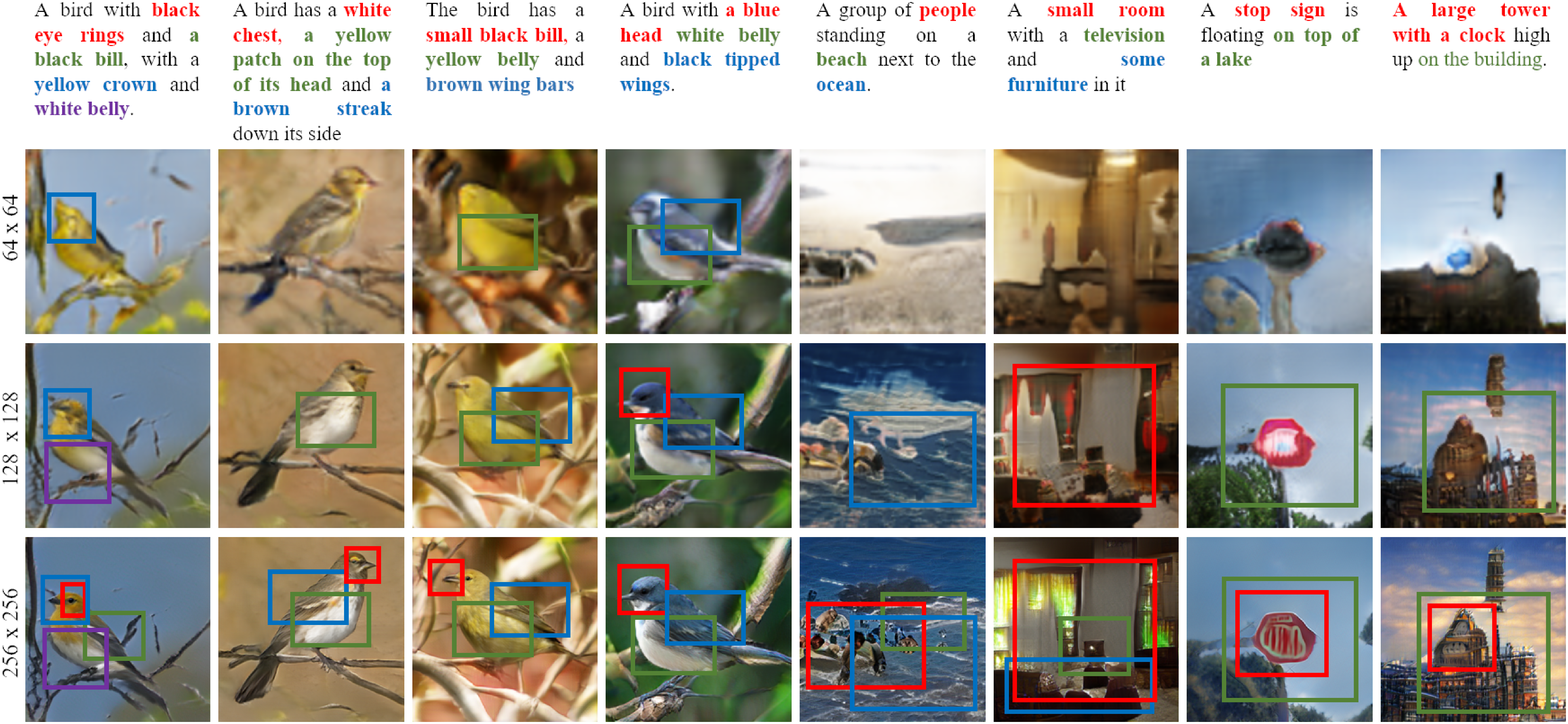}}
\end{figure}
\vskip -4mm
\centerline{\footnotesize\begin{tabular}{c} Fig.\ 5.\ Images of different stages generated by our FF-GAN on CUB and COCO datasets.
\end{tabular}}
\vskip -6mm
\begin{multicols}{2}

Observing the visual results on the CUB-200 dataset in Fig.4 (a), our method provides more authentic results than other models. Benefitting from the FF-Block, our model is able to take full advantage of the textual descriptions and mine more information ignored by other models. For instance, as can be seen in the 4th column of Fig.4(a), the given text ``this small bird has a yellow breast, brown crown, and white superciliary'', the image generated by our method contains all attributes, especially the attribute ``white superciliary'' which is neglected by others. In the 5th column, the detailed attribute ``short, roundish'' in the text is perfectly represented by our generated image. However, Attn-GAN$^{[5]}$ and DM-GAN$^{[35]}$ failed to reflect the ``roundish'' and generate a slender bird, which may be affected by the distribution of the training dataset. We could also observe some weakness of these methods in some cases. DM-GAN$^{[35]}$ generates an image in the 6th column that mismatched the attribution ``specked chest of brown and white'' and the bird generated by Attn-GAN$^{[5]}$ in the 6th column is not photo-realistic. The qualitative results indicate that our FF-GAN fuses the visual and linguistic features in a more effective and sufficient manner, which helps synthesize high-quality images that are semantically matched with the conditioned texts.

The results on the COCO dataset are shown in Fig.4(b). It should be pointed out that COCO is a much challenging dataset for the textual descriptions are more complicated and the corresponding images always contain multiple subjects with different backgrounds.
In the 1st column, cows generated by our method are clearly recognized and separated, while the ones synthesized by Attn-GAN$^{[5]}$ and DM-GAN$^{[35]}$ are mixed together and difficult to distinguish.  These qualitative results on COCO dataset demonstrate that our FF-GAN could synthesize visual-realistic pictures containing multiple complicated objects.

{\bf4.3.2\ Low-to-high resolution synthesis}

Fig.5 shows the images generated by our model in different stages.  It can be seen that the pictures generated in the initial stage are very fuzzy and lose a lot of textual information in the given text descriptions.
In the refinement progress, our model fully excavates the detailed information in the textual descriptions and fuses it into the visual features by FF-Block. So that the refined images can be matched with the input texts and generate more visual-realistic high-resolution images.
For example, as can be seen from the 1st column in Fig.5, the initial image with 64$\times$64 resolution is a mass of a yellow body, which merely capture the attribute ``yellow''. The refinement process helps to encode the attributes ``black bill'' ``white belly'' and generate high-resolution images containing missed linguistic information, and the quality of these refined images is significantly improved, with clear background and convincing details.

\end{multicols}
\vskip -2mm
\begin{figure}[H]
\includegraphics[width=0.92\textwidth]{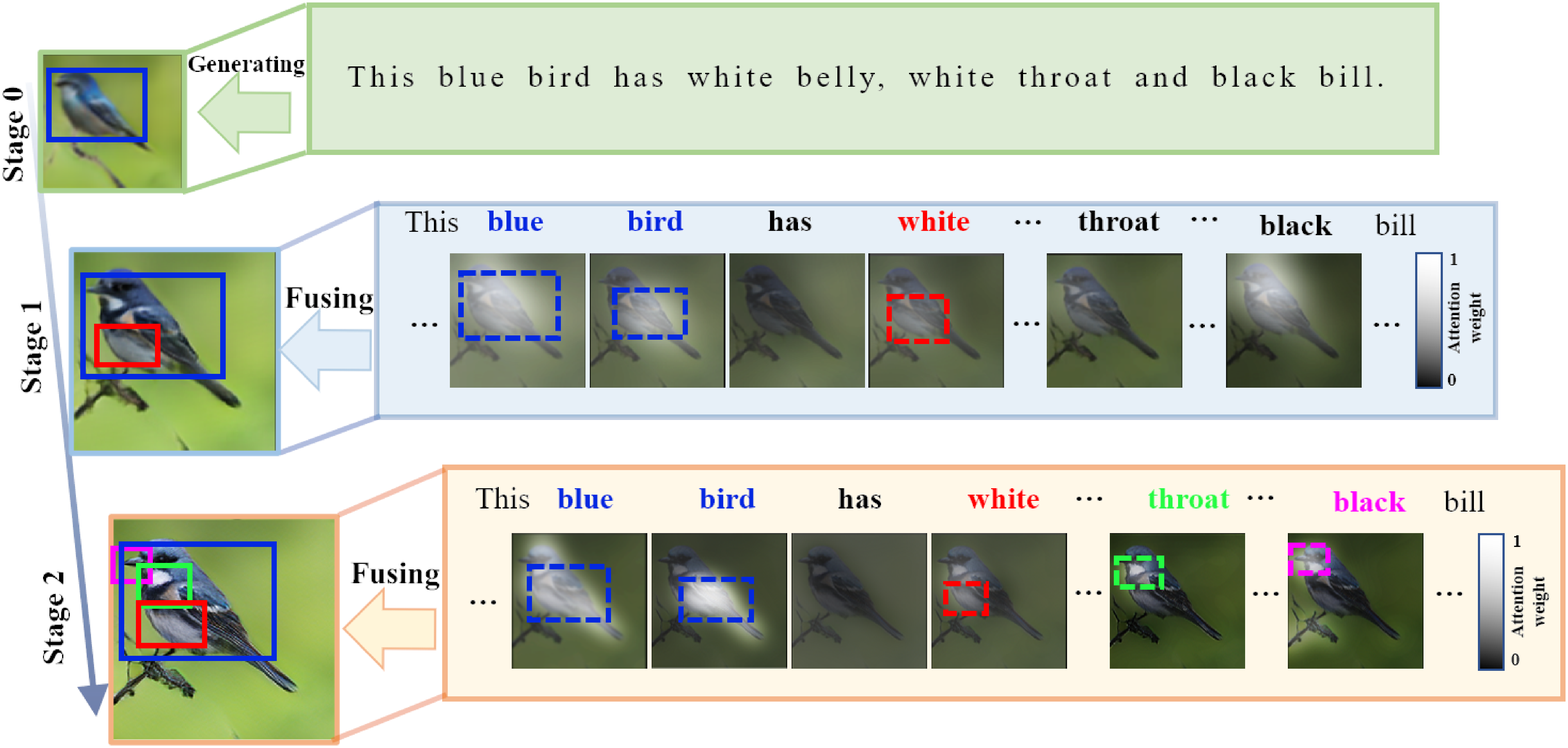}
\end{figure}
\vskip -2mm
\centerline{\footnotesize\begin{tabular}{l} Fig.\ 6.\ Visualization of the multi-stage generation process on the CUB-200. Our FF-GAN first generates a  low-resolution image by \\sentence-level features (stage 0), and exploits word-level features and additional sentence-level features to obtain a high-resolution \\\ image with more detailed information by two refinement stages (stage 1 and 2).
\end{tabular}}
\vskip -2mm

\begin{multicols}{2}

\vskip-4mm
\begin{figure}[H]
\centerline{\includegraphics[width=0.42\textwidth]{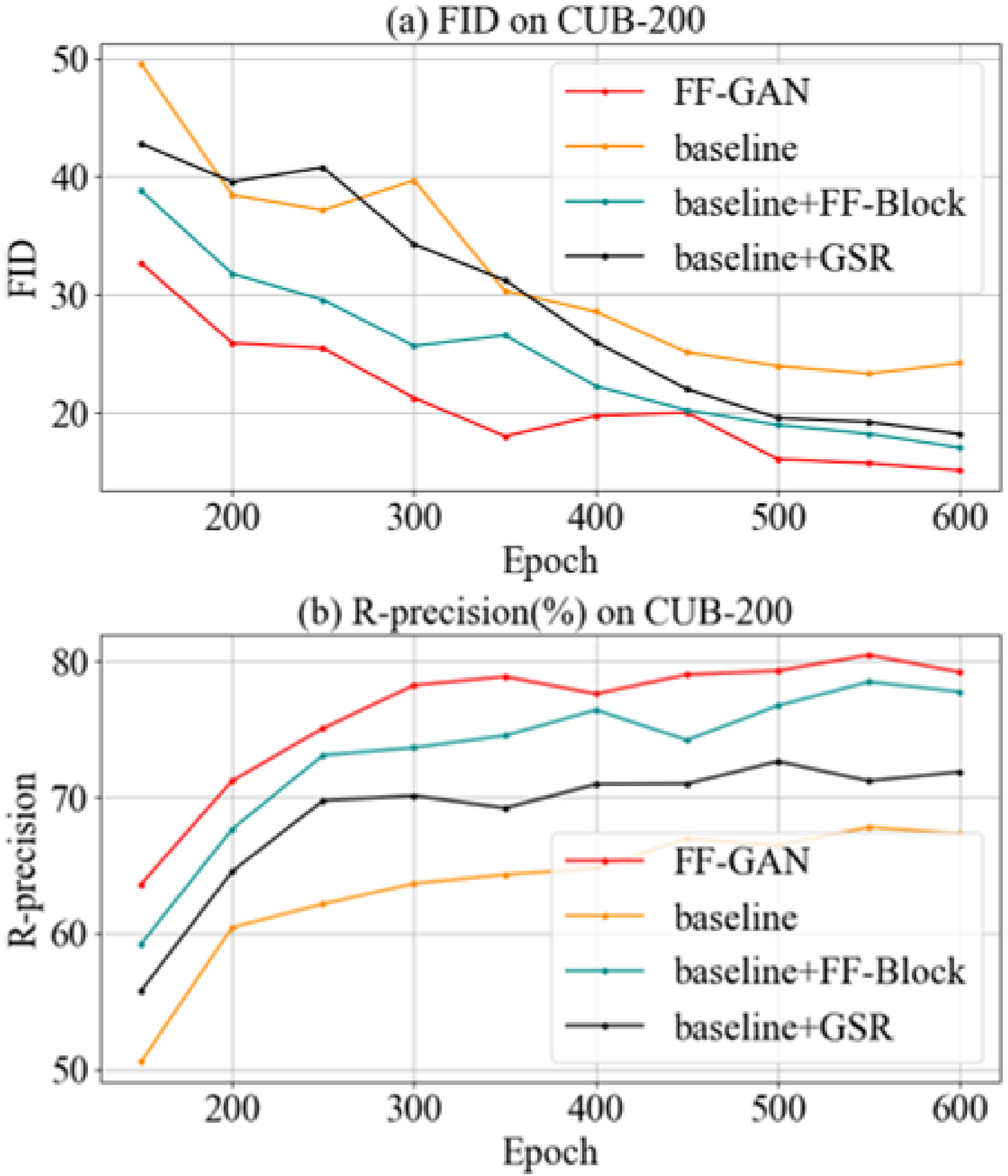}}
\end{figure}
\vskip -4mm
\centerline{\footnotesize\begin{tabular}{l} Fig.\ 7.\ FID and R-precision by our defined baseline and its \\ variants  at different epochs on CUB-200 test sets.
\end{tabular}}
\vskip 1mm

{\bf4.3.3\ Visual analysis on multi-stage refinement}

In order to understand how our proposed FF-GAN utilizes word-level linguistic features to refine the initial images, we compute the attention weights between word-level textual features and the image features, and visualize the intermediate attention maps of the word-level features in Fig.6. The 2nd and 3rd columns are the attention maps of some representative word at the first and second refinement stage respectively (stage 1 and stage 2 in Fig.6).

As can be seen from Fig.6, attention weights will be more allocated to words related to the generated images (the bright area in Fig.6), such as ``white'', ``throat'', and other attributes that describe body parts and colors. By contrast, the irrelevant words will be assigned with less attention and displayed in black on the attention maps, such as the word ``has''. In this way, by guiding the generator to focus on the most relevant words in the given texts during the refinement process, it helps the generative model refine the generated images with more fine-grained details.
At the same time, by comparing the attention maps in the 2nd column and the 3rd column, we could find that our model can refine the images of the previous stage to be more consistent with the given texts. For example, the attention map of the attribute ``black'' incorrectly highlights the whole head of the bird at stage 1, which results in generating the bad shape of the ``bill" in the generated image. In stage 2 (the 3rd column), the attention map of the word ``black'' highlights the area corresponding to the bill of the bird, which indicates that our model successfully learned this attribute at the 2nd refinement stage and refine the image finely, while the generated image also proves that our model learns the above detailed information correctly.

{\bf4.4\ Ablation studies}

{\bf4.4.1\ Ablation studies on key components}

We perform a series of ablation experiments on the CUB-200 dataset to evaluate the contribution of the key components of our FF-GAN. We define a baseline that removes FF-Block and GSR from FF-GAN, and some variants of baseline, such as ``baseline+FF-Block'' and ``baseline+GSR''. The performance of the baseline and its variants is reported in Fig.7.

\vskip-2mm
\begin{figure}[H]
\centerline{\includegraphics[width=0.42\textwidth]{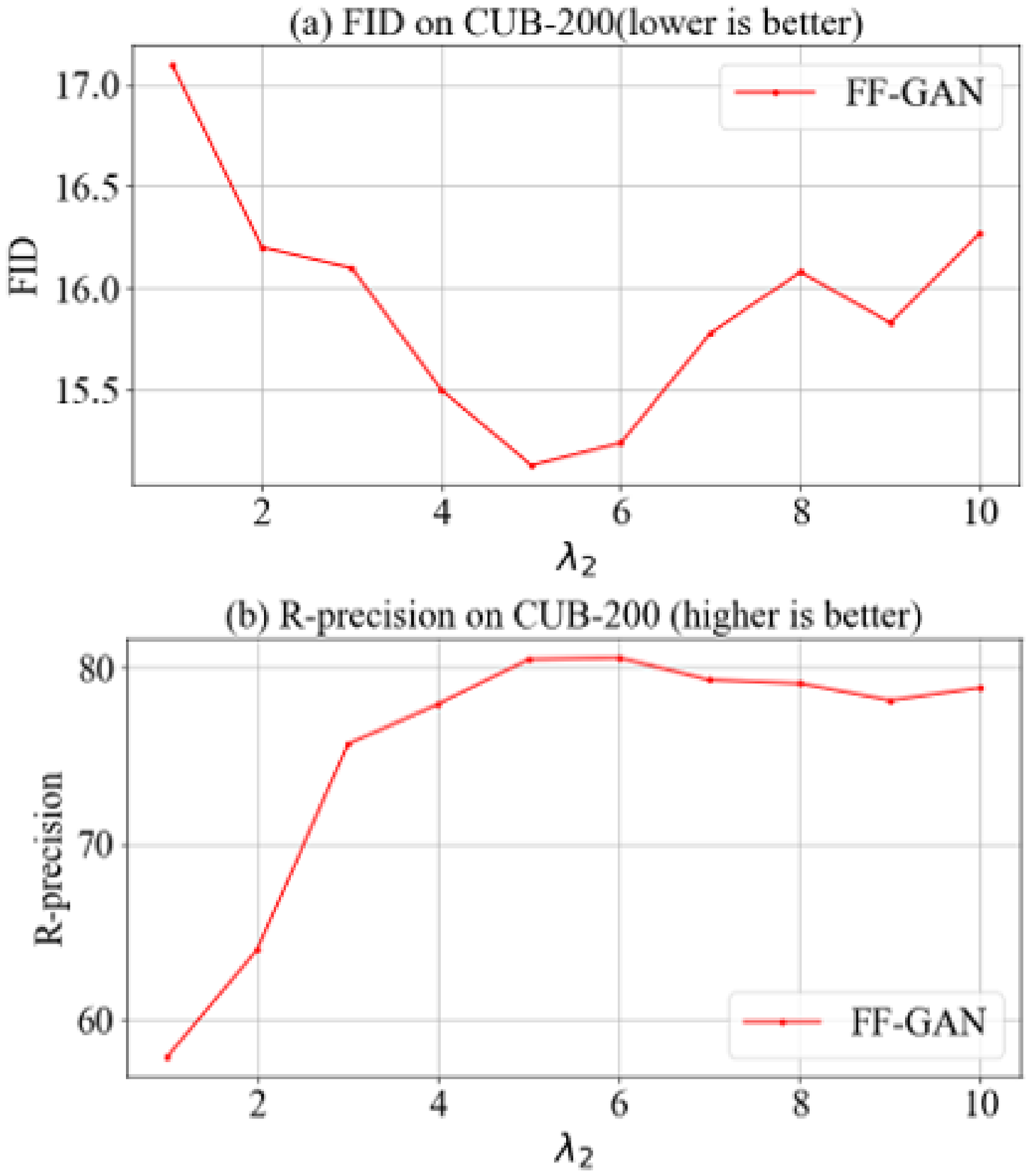}}
\end{figure}
\vskip -4mm
\centerline{\footnotesize\begin{tabular}{l} Fig.\ 8.\ FID and R-precision with different $\lambda_2$ settings on CUB-200 \\ test sets.
\end{tabular}}
\vskip 2mm

Comparing the yellow curve (baseline) and green curve (baseline + FF-Block) in Fig.7, we could draw a conclusion that FF-Block is able to ensure the textual-visual alignment and synthesize more visual-authentic pictures that more semantically match the given descriptions. FF-Block decreases the FID from 23.32 to 17.06 and increases R-precision by a large margin, from 67.82 to 78.52. The great promotion in R-precision indicates that FF-Block is able to fully integrate the fine-grained linguistic information into the visual features. Note that the R-precision of ``baseline + FF-Block'' is higher than R-precision of DM-GAN$^{[35]}$ and the FID exceeded most advanced methods, which indicate that the FF block is of great help in generating authentic and text-matched images.

The proposed GSR provides attentive sentence-level information to our FF-GAN in the refinement process. As can be seen from Fig.7, the baseline combined with GSR achieves 2.90 and 4.85 numerical improvements in FID and R-precision, respectively, which demonstrates that GSR is of great significance in improving the global semantic consistency between the synthesis images and the given texts. Moreover, the combination of GSR and FF-Block indicates that the two components can significantly reinforce the performance of the generative model and synthesize visual-realistic and semantically consistent images both locally and globally.

{\bf4.4.2\ Parameter studies}

The hyperparameters $\lambda_1$ and $\lambda_2$ in Eq.(10) are used to balance each loss, and the parameter setting makes no difference on the specific methods, so we adopted the typical setting following Attn-GAN$^{[5]}$ and DM-GAN$^{[35]}$. Specifically, the parameter $\lambda_1$ is used to balance the augmented data and real data, which is set to 1 by convention. The parameter $\lambda_2$ is used to control the text-image consistency and we set $\lambda_2$ to 5 on CUB-200 and 50 on COCO since $\lambda_2$ increases as the dataset becomes more complex.

To further verify the effectiveness of our parameter setting, we also conduct a series of verification experiments. Taking $\lambda_2$ on the CUB-200 dataset as an example, we set $\lambda_1$  to 1 by convention and set $\lambda_2$ from 1 to 10, with an interval of 1, and compare the results of FID  and R-precision.  As can be seen from Fig.8, the FID score (lower is better) decreases first and then increases as the $\lambda_2$ becomes larger, while the R-precision (higher is better) is opposite. Setting $\lambda_2$ to 5 can achieve the minimal FID score and maximal R-precision, where our model achieves the best performance. To sum up, the above experiments prove the effectiveness of our parameter setting.

\begin{center}{\large\bf V.\ Conclusions
}\end{center}

In this paper, we develop a novel Fine-grained text-image Fusion based Generative Adversarial Networks (FF-GAN) to synthesize images conditioned on textual descriptions. We design an effective Fine-grained text-image Fusion Block (FF-Block) to fully fuse visual and linguistic features, which helps synthesize more realistic and semantic consistent images based on the given linguistic descriptions. A Global Semantic Refinement (GSR) is employed to strengthen the semantic consistency of texts and pictures from a global semantic perspective. Both qualitative and quantitative experiments on two real-world benchmark datasets indicate that our proposed FF-GAN is superior to other advanced approaches in synthesizing high-quality visual-authentic images that match the texts.\\
\textbf{Acknowledgments} \ This work was supported by the National Natural Science Foundation of China under Grant No U21A20470, 62172136, U1936217. Key Research and Technology Development Projects of Anhui Province (no. 202004a5020043).



\RE

\footnotesize\rm
\REF{[1]} Perez E, Strub F, De Vries H, et al. Film: Visual reasoning with a general conditioning layer[C] //Proceedings of the AAAI Conference on Artificial Intelligence. 2018: 3942-3951

\REF{[2]} Reed S, Akata Z, Yan X, et al. Generative adversarial text to image synthesis[C]//International conference on machine learning. PMLR, 2016: 1060-1069.
\REF{[3]} Cheng J, Wu F, Tian Y, et al. RiFeGAN2: Rich Feature Generation for Text-to-Image Synthesis from Constrained Prior Knowledge[J]. IEEE Transactions on Circuits and Systems for Video Technology, 2021:5187 - 5200

\REF{[4]} Zhang H, Xu T, Li H, et al. Stackgan: Text to photo-realistic image synthesis with stacked generative adversarial networks[C]//Proceedings of the IEEE international conference on computer vision. 2017: 5907-5915.

\REF{[5]} Xu T, Zhang P, Huang Q, et al. Attngan: Fine-grained text to image generation with attentional generative adversarial networks[C]//Proceedings of the IEEE conference on computer vision and pattern recognition. 2018: 1316-1324.
\REF{[6]} Dumoulin V, Perez E, Schucher N, et al. Feature-wise transformations: A simple and surprisingly effective family of conditioning mechanisms[J]. Distill, 2018, 3(7): 1-11.

\REF{[7]} Liu B, Song K, Zhu Y, et al. TIME: Text and Image Mutual-Translation Adversarial Networks[C]//Proceedings of the AAAI Conference on Artificial Intelligence. 2021, 35 (3): 2082-2090.

\REF{[8]} Wang Y, Zhang W, Wu L, et al. Iterative views agreement: an iterative low-rank based structured optimization method to multi-view spectral clustering[C]//Proceedings of the Twenty-Fifth International Joint Conference on Artificial Intelligence. 2016: 2153-2159.
\REF{[9]} Li B, Qi X, Lukasiewicz T, et al. Controllable text-to-image generation[J]. Advances in Neural Information Processing Systems, 2019: 2065-2075.
\REF{[10]} Mao F, Ma B, Chang H, et al. MS-GAN: Text to Image Synthesis with Attention-Modulated Generators and Similarity-aware Discriminators[C]//The British Machine Vision Conference. 2019: 568-576.
\REF{[11]} Reed S, Akata Z, Mohan S, et al. Learning what and where to draw[C]//Proceedings of the 30th International Conference on Neural Information Processing Systems. 2016: 217-225.

\REF{[12]} Nguyen A, Clune J, Bengio Y, et al. Plug and play generative networks: Conditional iterative generation of images in latent space[C]//Proceedings of the IEEE conference on computer vision and pattern recognition. 2017: 4467-4477.

\REF{[13]} Zhang Z, Xie Y, Yang L. Photographic text-to-image synthesis with a hierarchically-nested adversarial network[C]//Proceedings of the IEEE conference on computer vision and pattern recognition. 2018: 6199-6208.
\REF{[14]}Wang Y. Survey on deep multi-modal data analytics: Collaboration, rivalry, and fusion[J]. ACM Transactions on Multimedia Computing, Communications, and Applications (TOMM), 2021, 17(1s): 1-25.

\REF{[15]} El-Nouby A, Sharma S, Schulz H, et al. Tell, draw, and repeat: Generating and modifying images based on continual linguistic instruction[C]//Proceedings of the IEEE/CVF International Conference on Computer Vision. 2019: 10304-10312.

\REF{[16]} Qian B, Wang Y, Hong R, et al. Rethinking Data-Free Quantization as a Zero-Sum Game[C]//Proceedings of the AAAI Conference on Artificial Intelligence. 2023: 1-8.

\REF{[17]} Park T, Liu M Y, Wang T C, et al. Semantic image synthesis with spatially-adaptive normalization[C]//Proceedings of the IEEE/CVF conference on computer vision and pattern recognition. 2019: 2337-2346.

\REF{[18]} Miyato T, Koyama M. cGANs with projection discriminator[J]. arXiv preprint arXiv:1802.05637, 2018.

\REF{[19]} Qian B, Wang Y, Yin H, et al. Switchable Online Knowledge Distillation[C]//European Conference on Computer Vision (11) 2022: 449-466.

\REF{[20]} Huang W, Da Xu R Y, Oppermann I. Realistic image generation using region-phrase attention[C]//Asian Conference on Machine Learning. PMLR, 2019: 284-299.
\REF{[21]} Heusel M, Ramsauer H, Unterthiner T, et al. Gans trained by a two time-scale update rule converge to a local nash equilibrium[J]. Advances in neural information processing systems, 2017: 6629-6640.

\REF{[22]} Lin T Y, Maire M, Belongie S, et al. Microsoft coco: Common objects in context[C]//European conference on computer vision. Springer, Cham, 2014: 740-755.

\REF{[23]} Wah C, Branson S, Welinder P, et al. The caltech-ucsd birds-200-2011 dataset[J]. 2011.

\REF{[24]} Schuster M, Paliwal K K. Bidirectional recurrent neural networks[J]. IEEE transactions on Signal Processing, 1997, 45 (11): 2673-2681.

\REF{[25]} Wu L, Wang Y, Shao L. Cycle-consistent deep generative hashing for cross-modal retrieval[J]. IEEE Transactions on Image Processing, 2018, 28(4): 1602-1612.

\REF{[26]} Mikolov T, Karafiát M, Burget L, et al. Recurrent neural network based language model[C]//Interspeech. 2010, 2 (3): 1045-1048.

\REF{[27]} Zhu J Y, Park T, Isola P, et al. Unpaired image-to-image translation using cycle-consistent adversarial networks[C]//Proceedings of the IEEE international conference on computer vision. 2017: 2223-2232.

\REF{[28]} Zhang Z, Schomaker L. DTGAN: Dual attention generative adversarial networks for text-to-image generation[C]//2021 International Joint Conference on Neural Networks  (IJCNN). IEEE, 2021: 1-8.
\REF{[29]}Liu H, Wang Y, Wang M, et al. Delving Globally into Texture and Structure for Image Inpainting[C]//Proceedings of the 30th ACM International Conference on Multimedia. 2022: 1270-1278.
\REF{[30]} Vaswani A, Shazeer N, Parmar N, et al. Attention is all you need[C]//Proceedings of the 31st International Conference on Neural Information Processing Systems. 2017: 6000-6010.

\REF{[31]} He S, Liao W, Yang M Y, et al. Context-aware layout to image generation with enhanced object appearance[C]//Proceedings of the IEEE/CVF Conference on Computer Vision and Pattern Recognition. 2021: 15049-15058.
\REF{[32]} Goodfellow I, Pouget-Abadie J, Mirza M, et al. Generative adversarial nets[J]. Advances in neural information processing systems, 2014: 2672–2680.

\REF{[33]} Qiao T, Zhang J, Xu D, et al. Mirrorgan: Learning text-to-image generation by redescription[C]//Proceedings of the IEEE/CVF Conference on Computer Vision and Pattern Recognition. 2019: 1505-1514.

\REF{[34]} Yin G, Liu B, Sheng L, et al. Semantics disentangling for text-to-image generation[C]//Proceedings of the IEEE/CVF conference on computer vision and pattern recognition. 2019: 2327-2336.

\REF{[35]} Zhu M, Pan P, Chen W, et al. Dm-gan: Dynamic memory generative adversarial networks for text-to-image synthesis[C]//Proceedings of the IEEE/CVF Conference on Computer Vision and Pattern Recognition. 2019: 5802-5810.

\REF{[36]} Cheng J, Wu F, Tian Y, et al. RiFeGAN: Rich feature generation for text-to-image synthesis from prior knowledge[C]//Proceedings of the IEEE/CVF Conference on Computer Vision and Pattern Recognition. 2020: 10911-10920.
\REF{[37]} Wang Y, Peng J, Wang H, et al. Progressive learning with multi-scale attention network for cross-domain vehicle re-identification[J]. Science China Information Sciences, 2022, 65(6): 1-15.

\REF{[38]} Tan H, Liu X, Liu M, et al. KT-GAN: knowledge-transfer generative adversarial network for text-to-image synthesis[J]. IEEE Transactions on Image Processing, 2020, 30: 1275-1290.

\REF{[39]} Szegedy C, Vanhoucke V, Ioffe S, et al. Rethinking the inception architecture for computer vision[C]//Proceedings of the IEEE conference on computer vision and pattern recognition. 2016: 2818-2826.

\REF{[40]} Zhang H, Xu T, Li H, et al. Stackgan++: Realistic image synthesis with stacked generative adversarial networks[J]. IEEE transactions on pattern analysis and machine intelligence, 2018, 41 (8): 1947-1962.
\REF{[41]} Qian B, Wang Y, Hong R, et al. Diversifying inference path selection: Moving-mobile-network for landmark recognition[J]. IEEE Transactions on Image Processing, 2021, 30: 4894-4904.

\REF{[42]} Hochreiter S, Schmidhuber J. Long short-term memory[J]. Neural computation, 1997, 9 (8): 1735-1780.
\REF{[43]} De Vries H, Strub F, Mary J, et al. Modulating early visual processing by language[C]//Proceedings of the 31st International Conference on Neural Information Processing Systems. 2017: 6597-6607.
\REF{[44]} Tao M, Tang H, Wu F, et al. DF-GAN: A Simple and Effective Baseline for Text-to-Image Synthesis[C]//Proceedings of the IEEE/CVF Conference on Computer Vision and Pattern Recognition. 2022: 16515-16525.

\REF{[45]} Huang X, Belongie S. Arbitrary style transfer in real-time with adaptive instance normalization[C]//Proceedings of the IEEE international conference on computer vision. 2017: 1501-1510.

\REF{[46]} Ruan S, Zhang Y, Zhang K, et al. Dae-gan: Dynamic aspect-aware gan for text-to-image synthesis[C]//Proceedings of the IEEE/CVF International Conference on Computer Vision. 2021: 13960-13969.

\REF{[47]} Li B, Qi X, Lukasiewicz T, et al. Manigan: Text-guided image manipulation[C]//Proceedings of the IEEE/CVF Conference on Computer Vision and Pattern Recognition. 2020: 7880-7889.

\vskip 4mm

\includegraphics[width=0.11\textwidth]{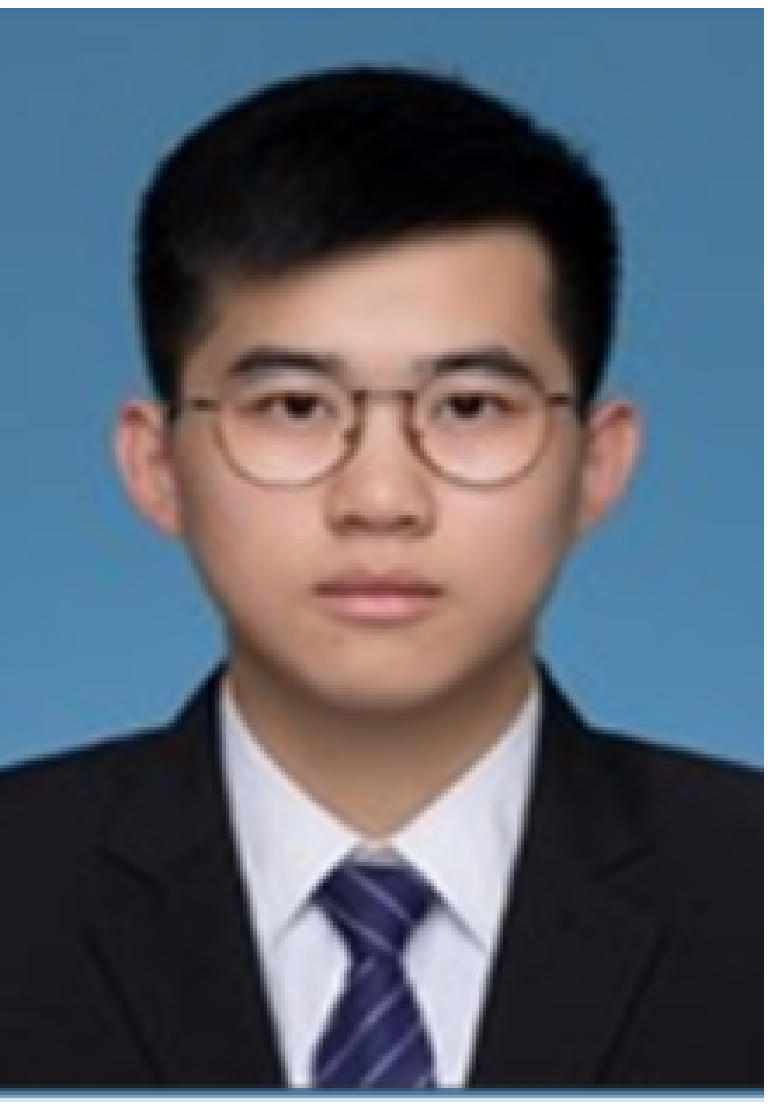}

\vskip -3cm

\hangindent=2.6cm \hangafter=-10 {\bf SUN Haoran
}\quad received the B.E. degree in 2020. He is currently pursuing the M.S. degree at Hefei University of Technology. His research interests include artificial intelligence and computer vision.   (Email:  haoran\_sun@mail.hfut.edu.cn)

\vskip 3cm

\includegraphics[width=0.11\textwidth]{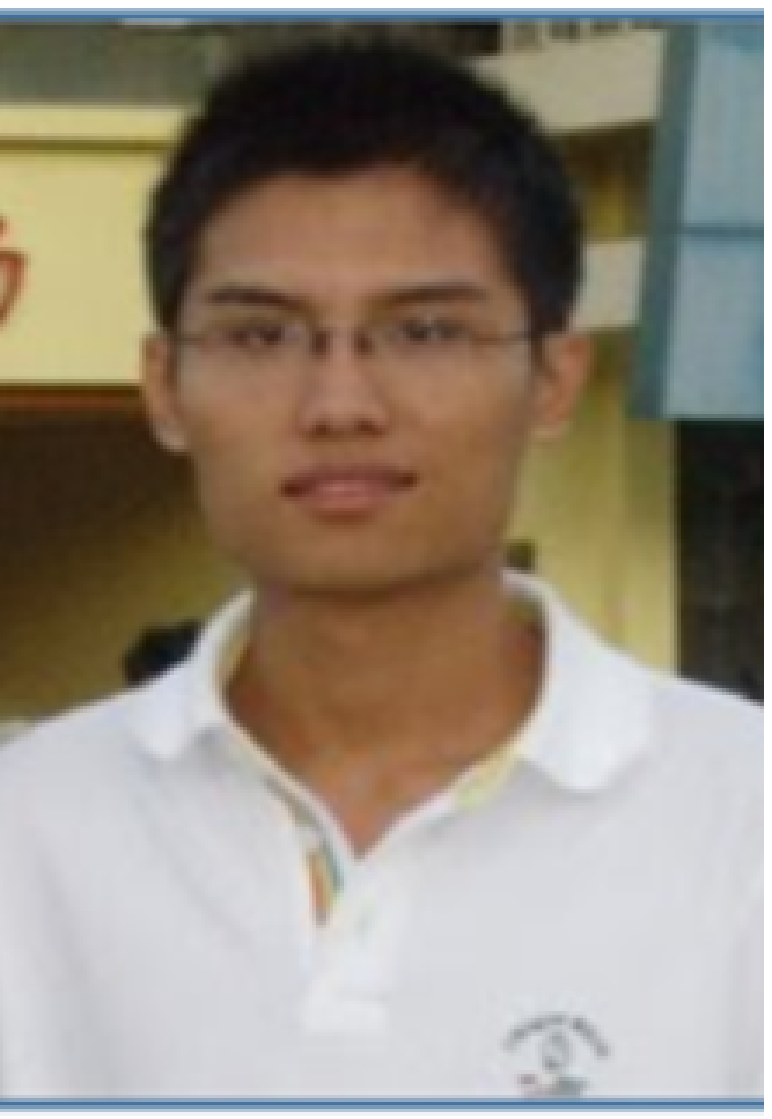}

 \vskip -3.0cm

\hangindent=2.6cm \hangafter=-10 {\bf WANG Yang
}\quad(corresponding author) earned a PhD degree (2015.09) from The University of New South Wales, Kensington, Australia. He is currently a Huangshan Professor and PhD supervisor at Hefei University of Technology, China. He has published 90 research papers (featured 7 ESI highly cited papers, with all of them to be among Top 1\% ) including two book chapters, most of which are (to be) appeared in the major venues, such as Artificial Intelligence (Elsevier), International Journal of Computer Vision (IJCV), IEEE TIP, IEEE TNNLS, IEEE TMM, ACM TOIS, Machine Learning (Springer), IEEE TKDE, IEEE TCYB, VLDB Journal, KDD, ECCV, IJCAI, AAAI, ACM SIGIR, ACM Multimedia, IEEE ICDM, ACM CIKM, SCIENCE CHINA Information Sciences etc. He currently serves as the Associate Editor of ACM Trans. Information systems. He was the winner of Best Research Paper Runner-up Award for PAKDD 2014, and was a program committee member for various leading conferences such as IJCAI, AAAI, CVPR, ECCV, EMNLP, ACM Multimedia, ACM Multimedia (Asia), ECMLPKDD, etc.   (Email:   yangwang@hfut.edu.cn)

\vskip 2cm

\includegraphics[width=0.11\textwidth]{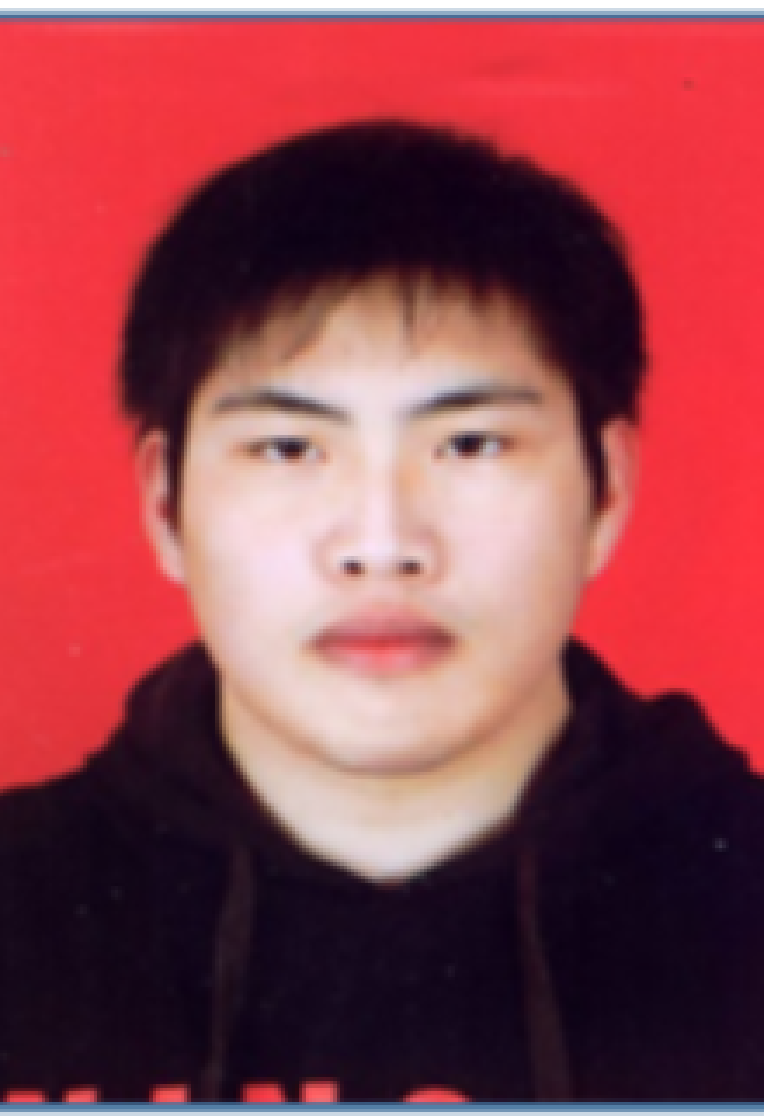}
\vskip -3cm

\hangindent=2.6cm \hangafter=-10 {\bf LIU Haipeng
}\quad received the B.E. degree in 2018. He is a Ph.D. candidate at the Hefei University of Technology, Hefei, China. His current research interests include computer vision, deep learning and image inpainting. (Email:   hpliu\_hfut@hotmail.com)

\vskip 2cm

\includegraphics[width=0.11\textwidth]{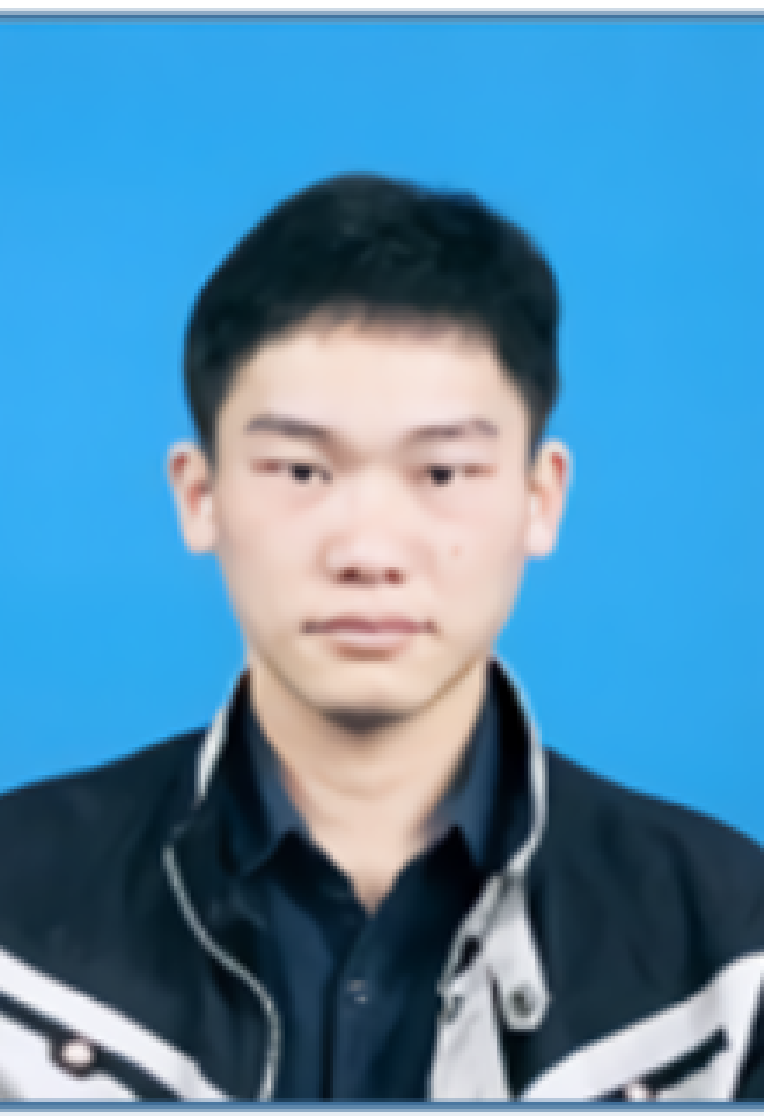}

\vskip -3cm

\hangindent=2.6cm \hangafter=-10 {\bf QIAN Biao
}\quad  is a Ph.D. candidate and received the B.E. degree in 2017 at the Hefei University of Technology, Hefei, China. His current research interests include computer vision, deep learning and neural network compression and acceleration. (Email: hfutqian@gmail.com)









\end{multicols}
\end{document}